\pdfoutput=1
\documentclass[11pt]{article}

\usepackage[final]{acl}

\usepackage{times}
\usepackage{latexsym}

\usepackage[T1]{fontenc}
\usepackage[utf8]{inputenc}

\usepackage{microtype}

\usepackage{inconsolata}

\usepackage{graphicx}
\usepackage{amsmath}
\usepackage{listings}

\colorlet{punct}{red!60!black}
\definecolor{background}{HTML}{EEEEEE}
\definecolor{delim}{RGB}{20,105,176}
\colorlet{numb}{magenta!60!black}

\lstdefinelanguage{json}{
    basicstyle=\scriptsize\ttfamily,
    stepnumber=1,
    numbersep=8pt,
    showstringspaces=false,
    breaklines=true,
    breakindent=0pt,
    frame=lines,
    backgroundcolor=\color{background},
}
\title{An Approach to Build Zero-Shot Slot-Filling System \\ 
for Industry-Grade Conversational Assistants}

\author{
  G P Shrivatsa Bhargav$^\dagger$$^\ast$,~~Sumit Neelam$^\dagger$$^\ast$,~~Udit Sharma$^\dagger$,~~Shajith Ikbal$^\dagger$, \\
  \textbf{Dheeraj Sreedhar$^\dagger$,~~Hima Karanam$^\dagger$$^\ast$,~~Sachindra Joshi$^\dagger$,~~Pankaj Dhoolia$^\dagger$,} \\
  \textbf{Dinesh Garg$^\dagger$,~~Kyle Croutwater$^\ddagger$$^\ast$,~~Haode Qi$^\ddagger$,~~Eric Wayne$^\ddagger$,~~J William Murdock$^\ddagger$} \\
  {\fontsize{1.5}{4}\selectfont ~} \\
  {\large{\textbf{$^\dagger$~IBM Research}}}~~~~~~~~~{\large{\textbf{$^\ddagger$~IBM Watson}}} \\
  $^\ast$~\{gpshri27, sumit.neelam, hkaranam\}@in.ibm.com,~~kylec@us.ibm.com
  {\Large ~}
  }

\begin{document}
\maketitle

\begin{abstract}
We present an approach to build Large Language Model (LLM) based slot-filling system to perform Dialogue State Tracking in conversational assistants serving across a wide variety of industry-grade applications. Key requirements of this system include: 1) usage of smaller-sized models to meet low latency requirements and to enable convenient and cost-effective cloud and customer premise deployments, and 2) zero-shot capabilities to serve across a wide variety of domains, slot types and conversational scenarios. We adopt a fine-tuning approach where a pre-trained LLM is fine-tuned into a slot-filling model using task specific data. The fine-tuning data is prepared carefully to cover a wide variety of slot-filling task scenarios that the model is expected to face across various domains. We give details of the data preparation and model building process. We also give a detailed analysis of the results of our experimental
evaluations. Results show that our prescribed approach for slot-filling model building has resulted in 6.9\% relative improvement of F1 metric over the best baseline on a realistic benchmark, while at the same time reducing the latency by 57\%. More over, the data we prepared has helped improve F1 on an average by 4.2\% relative across various slot-types.

%We present an approach to build Large Language Model (LLM) based slot-filling system to perform Dialogue State Tracking in conversational assistants serving across a wide variety of industry-grade applications. Key requirements of this system include: usage of smaller-sized models to enable convenient and cost-effective cloud and customer premise deployments, and zero-shot inference capabilities to serve across a wide variety of domains, slot types and conversational scenarios. We adopt a fine-tuning approach where a pre-trained LLM is fine-tuned into a slot-filling model using task specific data. The fine-tuning data is prepared carefully to cover wide variety of slot-filling task scenarios that the model is expected to face across various domains. We give details of the data preparation and model building process. We also give a detailed analysis of the results of experiments conducted to evaluate our model across various feature requirement dimensions, in comparison to the baselines.
\end{abstract}

\section{Introduction}

Using task-oriented dialogue systems to build conversational assistants serving a wide range of domains is a long sought after goal of industries looking to enhance their applications with conversational capabilities. This has gained a renewed focus in recent times with the advent of Large Language Models (LLM). A key component in these systems is Dialog State Tracking (DST), to estimate the evolving goals of the users during the course of conversation. DST involves gathering relevant domain-specific information from the conversation so far, that could be used to interface with external Application Program Interfaces (API), to eventually generate appropriate responses to the users. Interfacing with the APIs is an effective mechanism to combine superior language capabilities of LLMs with the relevant domain-specific knowledge obtained through APIs, to build powerful applications. 
\begin{figure}[t!]
  \centering\includegraphics[keepaspectratio,width=0.48\textwidth]{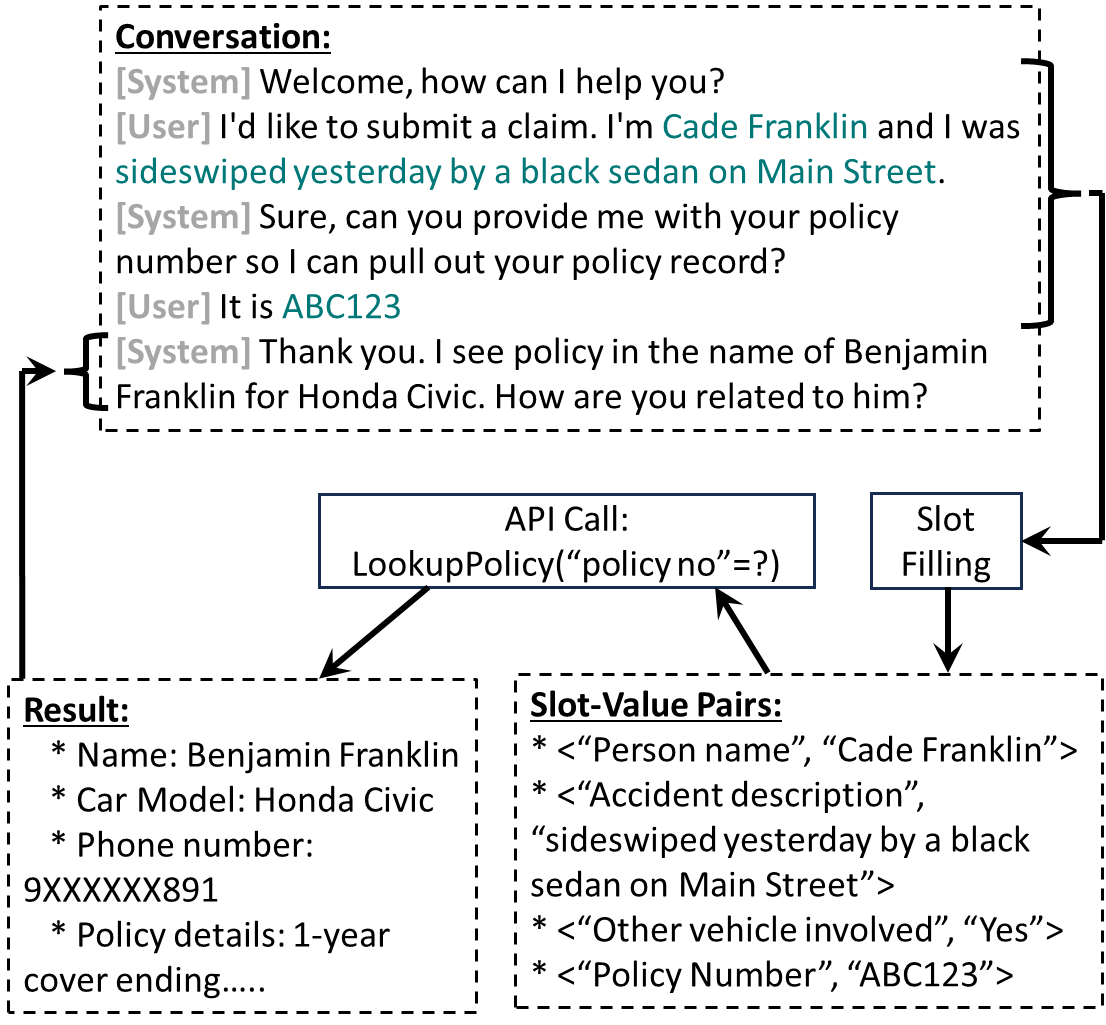}
  \caption{A simple illustration of how API calling enabled by Slot-Filling model is utilized in conversational assistants.
  }
  \label{fig:illus}
\end{figure}

Domain-specific user goals are typically represented as a set of slot-value pairs, i.e., a set of parameters and their corresponding values needed to execute functions represented by the APIs, as illustrated in Figure \ref{fig:illus}. Estimation of domain-specific slot-value pairs from the conversational history is termed as Slot-Filling task.
In this paper, we describe an approach to build 
LLM based slot-filling systems that are generic, robust and appropriate enough to serve industry-grade applications.
%generic and robust slot filling systems suitable to serve industry-grade applications. 
Some of the key feature requirements of such systems include: 1) Usage of smaller models (of around 10B parameters) that are ideal for training and deployment on cloud or on premises, unlike massive models such as ChatGPT that typically have restrictions in terms of slow response time, prohibitive inference costs, and data privacy concerns. 2) Ability to handle a wide variety of slots/entities/scenarios 
%(typically needed to build robust applications that can serve across a wide variety of domains) such as categorical slots, name splitting, address parsing, relation extraction, multiple slots expressed in single user turns, conversations and slot descriptions containing markdowns, HTMLs, etc., slot value rewrite during the conversation, various ids such as email id, customer id etc. 
typically needed to build robust applications that can serve across a wide variety of domains.
3) Zero-shot capabilities to work well on domains and scenarios not seen during the training, by relying just on the instructions provided through prompts during inference.

We adopt a fine-tuning approach to build slot-filling model, where a task-specific dataset is prepared to fine-tune a pre-trained LLM into a slot-filling model, which is further used in a zero-shot fashion to perform slot-filling. The fine-tuning data is prepared from a mix of SGD dataset~\cite{sgd} and a set of human-generated datasets incorporating basic features that we believe are needed to cover a wide range of scenarios that our model is expected to face such as: a) a variety of domains, b) a variety of slot-value extraction scenarios such as multiple slots expressed in a single user turn, slot-value rewrite during the conversation, and relation extraction, 
%and conversations and slot descriptions containing markdowns, HTMLs, etc, 
and c) a variety of slot-value types such as categorical slots, long slot values, address parsing, name splitting, and various ids including email id, customer id, etc. 

After 
%a formal definition of the slot-filling problem in Section \ref{sec:prob-def} and 
giving
a description of our model building strategy in Section 
%\ref{sec:train}, 
\ref{sec:model},
we describe in detail the fine-tuning data in Section \ref{sec:data}. 
%Section \ref{sec:results} gives results of experimental evaluation demonstrating the effectiveness of our slot-filling model.
Section \ref{sec:results} demonstrates the effectiveness of our approach through results of experimental evaluation conducted on three slot-filling models, built by fine-tuning two publicly available LLMs \textit{Mistral} \cite{jiang2023mistral} and \textit{Flan-T5-XL} \cite{chung2022scaling} and our proprietary model %referred to in this paper as \textit{inhouse-13b}.
\textit{granite.13b.v2} \cite{ibmgranite}.
Our contributions in this paper are: % as follows:
\begin{itemize}
    \item A set of slot-filling evaluation benchmarks spanning across a list of basic features that we believe would encapsulate the real world scenarios that the slot-filling models are likely to operate in.
    \item Description of how to construct the set of datasets that teach LLM based models on handling a variety of slot types in a  variety of conversational scenarios.
    \item A thorough experimental evaluation of our approach, illustrated through fine-tuning of three LLMs - two publicly available LLMs namely \textit{Mistral} and \textit{Flan-T5-XL} and our proprietary LLM \textit{granite.13b.v2} - demonstrating that the additional data prepared (to teach the model on variety of slot types and scenarios) is able to enable even smaller-sized models (of around 7-13B parameters) to achieve slot-filling accuracy metrics comparable to or better than that of massive models such as ChatGPT and PaLM.
    %\item Training and evaluation of the models on SGD to show that these models are not robust and are limited to only certain types of slots.
    %\item Training and evaluation of models on SGD + our data, showing that even small models having around 7-13B parameters can achieve accuracy comparable to or better than very large and capable models such as ChatGPT and PaLM.
\end{itemize}

\section{Related Works}
Slot-filling for dialogue state tracking is an active area of research with open datasets built specifically for the task such as Multiwoz ~\cite{budzianowski-etal-2018-multiwoz} Schema Guided Dialogue (SGD) and SGD-X~\cite{sgd,DBLP:journals/corr/abs-2110-06800}. In recent years zero-shot slot-filling using large language models has gained momentum, with specific focus on improving API calling abilities~\cite{qin2023tool,tang2023toolalpaca,qin2023toolllm,chatgpt-plugin}.

Early works on slot-filling have looked at learning representations for the slots from the descriptions and examples in the conversations of training data~\cite{DBLP:journals/corr/BapnaTHH17aa,qin-etal-2019-stack,lee2018zeroshot,9033997,shah-etal-2019-robust}.
Later on works have explored building custom architectures towards tackling dialogue state tracking like GNN, RNN, or Bi-LSTM networks~\cite{guo-etal-2022-beyond,10.1145/3511808.3557581,feng2022dynamic}, and also considered specialized network such as Dialogue State Distillation Network (DSDN)~\cite{xu2023dialogue}.

Large focus in recent times is on using large language models to achieve the slot-filling and dialogue state tracking. One direction of work use prompting or prompt tuning methods or selection of few shot examples for prompting~\cite{feng-etal-2023-towards,venkateswaran-etal-2023-district,su-etal-2023-scalable,ma2023parameterefficient,qixiang-etal-2022-exploiting}.
Several recent works have taken the approach of fine tuning large language models using the existing open source data to perform slot-filling task~\cite{shi-etal-2023-adaptive,zhang2023turnlevel,10.1145/3534678.3539385,wang-xin-2022-stop,zhao2022descriptiondriven,Gupta_2022}. Our fine-tuning approach to build slot-filling model is similar to ~\cite{zhao2022descriptiondriven}, but our focus is beyond academic benchmarks, towards filling the gaps to make it industry grade. In that respect we curate additional training data to address the gaps and use that along with the SGD data to fine tune a few open source LLMs and our proprietary \textit{granite.13b.v2} %inhouse-13b 
model, to show the effectiveness of our approach.

\section{Slot-Filling Model Building}
\label{sec:model}
%\section{Model Training Methodology}
%\label{sec:train}
We give a description of our model building approach, after defining the slot-filling problem.

\subsection{Slot-Filling Problem}
%The goal is to extract a set of slot-value pairs expressed in a conversation between an user and an agent. 
Let $C_t$ denote a conversation of $t$ turns: 
\[ C_t = \{U_1, A_1, U_2, A_2, ..., U_t\} \] 
where $U_i$ and $A_i$ are respectively the user and agent utterances, and let $S$ denote specifications of a set of $J$ slots:
\[ S = \{S_1, S_2, ..., S_J\} \]
where each slot specification $S_j$ in turn is a tuple
$S_j = (S^{\text{id}}_j, S^{\text{name}}_j, S^{\text{desc}}_j$),
with $S^{\text{id}}_j$ denoting unique alphanumeric identifier for the slot, $S^{\text{name}}_j$ its name, and $S^{\text{desc}}_j$ its natural language description.
%$S_j=(S^{\text{id}}_j, S^{\text{desc}}_j)$
The goal of slot-filling task is to extract values for the slots as expressed in $C_t$ through mapping:
\begin{equation}
\label{equ:map}
    \mathcal{F}: (\mathcal{C}_t, \mathcal{S}) \rightarrow \{(S^{\text{id}}_1, V^t_1), (S^{\text{id}}_2, V^t_2), ..., (S^{\text{id}}_J, V^t_J)\}
\end{equation}
%where $B_t$ denote estimated slot-value pairs
%\[ B_t = \{(S^{\text{id}}_1, V^t_1), (S^{\text{id}}_2, V^t_2), ..., (S^{\text{id}}_J, V^t_J)\} \]
where 
$V^t_j$ denote value for slot id $S^{\text{id}}_j$ inferred from the conversation until turn $t$. 
%Note that $V^t_j$ and $V^{t+1}_j$ may be different as the user might have updated/corrected the information in utterance $U_{t+1}$.
Note that all $V^t_j$ are initialized to none at the start and a subset of them %their values 
are expected to get assigned/modified as per the revelation/updation of the corresponding values by the user at each turn $t$. %change at turn $t$ depending upon user adding/modifying the slot value at $U_{t}$.

\subsection{Model Building}
Our approach to build slot-filling model: fine-tune a pre-trained LLM to make it learn the mapping in (\ref{equ:map}).
%We believe the resulting model's ability to handle language understanding intricacies would further help improve slot-filling accuracy.
%We believe the resulting model for slot filling task would also benefit from its original ability to handle language understanding intricacies.
We believe such an approach would enable the model to utilize its language understanding capabilities towards effective slot-filling.
A key step in our approach is the preparation of fine-tuning data. We want the model to learn more generic mapping (\ref{equ:map}) from the examples given in the fine-tuning data, so it can perform slot-value extraction in a zero-shot manner across a wide variety of slot types in a wide variety of conversational scenarios. For this purpose, our data is shaped up to instruction fine-tune the LLMs, so the model can learn to adapt as per the instructions provided through prompts. Next, we give details of our fine-turning data format, and then the details of LLM fine-tuning.

\subsubsection{Fine-Tuning Data Format}
\label{sec:data-format}
Fine-tuning data contains two parts: 1) prompt that will be provided as input to the model, and 2) output that the model is expected to produce in a generative fashion.
Input prompt is made of the following fields:
\begin{itemize}
    \item Natural language (NL) description of the slot-filling task.
    \item Details of the set of slots that are expected to get filled during the conversation. In particular, we use slot ids $S^{\text{id}}_j$ and the NL descriptions of the slot $S^{\text{desc}}_j$.
    \item Conversation history $C_t$ until turn $t$, with details of user and agent utterances.
\end{itemize}
Note that we do not use slot names in the slots field of the input prompt. This is to force the model to learn to read slot descriptions instead of the slot names, that we found to be beneficial through experimental evaluations. Information about the specific set of slots to be filled, as provided through the slots field of the prompts, is expected to help the model adapt to the specific conversational scenario and slot type in a zero-shot fashion.
The output expected to be generated by the model is: 
\begin{itemize}
    \item Set of key value pairs where keys are slot ids $S^{\text{id}}_j$ specified in the input prompt and the values are the corresponding slot values.
\end{itemize}
Note that we process the output generated by the model further to ensure that the slot values are either substrings of the conversation in case of non-categorical slots or belong to a set of permitted values in case of categorical slots. Additionally during training, the total length of prompt plus output is restricted to a maximum limit of 1200 , 270 tokens respectively. In those cases where the total length exceed this length limit, early utterances of the conversational history $C_t$ are dropped out from the prompt to meet the length limit requirement. Details of the data preparation is given in Section~\ref{sec:data}.

\subsubsection{Fine-Tuning}
\label{sec:fine-tuning}
To illustrate our approach, we fine-tune three pre-trained LLMs - two publicly available LLMs namely \textit{Mistral} and \textit{Flan-T5-XL} and %one of 
our proprietary LLM %referred to as
\textit{granite.13b.v2}. %\textit{inhouse-13b}. 
In Section \ref{sec:results}, we give results of experimental evaluations conducted on three slot-filling models built from these three LLMs. \textit{Flan-T5-XL} (3B) is fully fine-tuned for 15,000 steps with batch size 2 and learning rate 1e-5. \textit{Mistral} (7B) is LoRA \cite{hu2021lora} fine-tuned for 15,000 steps with batch size 16 and learning rate 1e-4. \textit{granite.13b.v2} is also LoRA fine-tuned for 40,000 steps with batch size 6 and learning rate 1e-4. Same LoRA parameters (lora\_rank: 8, lora\_alpha: 32, lora\_dropout: 0.1) are used for both \textit{Mistral} and \textit{granite.13b.v2}. %\textit{inhouse-13b}. 
All experiments are performed using two A100 80GB GPUs. 

%We have performed full fine-tuning \cite{} in case of \textit{Mistral} and LoRA  \cite{hu2021lora} fine-tuning in case of \textit{Flan-T5-XL} and \textit{inhouse-13b}. To add model parameters... hyperparameters (model size, learning rate, batch size, lora rank, training time, hardware details which gpu and how many, average inference time).. and other details like training time and hardware requirement.

\if false
\subsection{Data Format}
We prepare fine-tuning data in a manner where input to the model together with any other additional information that we might want to add is provided as a prompt to the LLM, so it can produce the expected output in a generative fashion.
\subsubsection*{Input}
\begin{itemize}
    \item Prompt consists of the slot library and the conversation
    \item We use slot IDs instead of the slot names to force the model to learn to read the slot description instead of relying only on the slot name. This will help in zero-shot generalization to unseen domains.
\end{itemize}

\subsubsection*{Output}
\begin{itemize}
    \item We train the model to output key value pairs where the keys are the slot names and the values are slot values.
    \item For non-categorical slots, we ensure that the slot values are sub-strings of the conversation. 
    \item For categorical slots, we ensure that the slot values belong to the set of permitted values
\end{itemize}

\subsection{Training}
\begin{itemize}
    \item Methods: full fine-tuning, LORA, prompt tuning, etc
    \item Mention the models used. 
    \item Write the hyper-parameters, training time, hardware requirement, etc in the appendix or here?
\end{itemize}
\fi

%\section{Creating New Datasets, Training and Evaluating models}
%\label{sec:data}
\section{Data Preparation}
\label{sec:data}

%What do we call this section?

%In this section we outline the process we used to create training and evaluation data that cover various types of slots and conversation scenarios that frequently occur in the real world.
In this section, we outline the details of preparation of our fine-tuning data and evaluation data. The goal is to expose the slot-filling model to a few fundamentally distinct slot types during fine-tuning, with the hope that the model would then be able to extrapolate to a wide variety of slot types in a wide variety of conversational scenarios, typically seen in real world applications.
The evaluation data is divided as per the slot types and scenarios so we can do a category-wise model performance analysis as well.

Our dataset is made from two sources: 1) data prepared from a publicly available dataset namely SGD \cite{sgd} and 2) the data that we prepared/curated. 
Note that the data we prepared is to cover those slot types and scenarios that are not covered by the SGD and in some cases to also augment the data already prepared from SGD. The data is prepared as per the format described in Section \ref{sec:data-format}. %In case of SGD, the original data is reshaped into the required format. 
The datasets prepared on different categories of slot types and scenarios are divided further into train, validation and test sets for their corresponding role in fine-tuning and evaluation of the models. Table \ref{tab:data-example} shows data samples for a few specific slot types, that we describe next.
Sizes of train, validation and test splits of different components are also given in the first column of Table \ref{tab:data-example}.

\begin{table*}[t!]
\small
\centering
\begin{center}
\begin{tabular}{|p{1.35cm}|p{5.35cm}|p{5.5cm}|p{2.7cm}|}
\hline
\textbf{Slot Type} & \textbf{Slot Library} & \textbf{Conversation} & \textbf{Output} \\
\hline
\textbf{Multi-slot} \newline (tr:1056, v:140, ts:286) & 
Slot-5: City where bus is going to \newline
Slot-182: Number of travelers for journey. Allowed values ("1", "2", "3", "4", "5") \newline
Slot-53: Date of bus leaving for journey \newline
Slot-57: Time of bus leaving for journey \newline
Slot-24: City where bus is leaving from
& 
\textcolor{gray}{[User]} I need to book \textcolor{teal}{4} tickets for bus leaving from \textcolor{teal}{Fresno} to \textcolor{teal}{long beach} on \textcolor{teal}{March 10th} at \textcolor{teal}{1:40 pm}.
&Slot-5:[`long beach'], Slot-182:[`4'], \newline Slot-53:[`March 10th'], Slot-57:[`1:40 pm'],\newline  Slot-24:[`Fresno']\\
\hline
\textbf{Long value} \newline (tr:280, v:0, ts:191)& 
Slot-34: id of an order \newline
Slot-28: cancellation reason
& 
\textcolor{gray}{[User]} I want to cancel my order \textcolor{teal}{8978JHG} as \textcolor{teal}{delivery time is too far away from what I anticipated} \newline
\textcolor{gray}{[System]} sure, cancelled your order with ID 8978JHG.
& Slot-34:[`8978JHG'], Slot-28:[`delivery time is too far away from what I anticipated'] \\
\hline
\textbf{Categorical} \newline (tr:17321, v:2500, ts:5058) & 
Slot-51: Name of the hair stylist/salon\newline  
Slot-0: Time of the appointment \newline
Slot-154: Date for the appointment \newline
Slot-63: Please confirm. Allowed values ("Yes, go ahead","No")
& 
\textcolor{gray}{[User]} I need a salon appointment.\newline 
\textcolor{gray}{[System]} Do you have a preferred salon? What date and time do you have in mind for the appointment? \newline
\textcolor{gray}{[User]} I like an appointment at \textcolor{teal}{Salon Revel} on \textcolor{teal}{the 1st} in the \textcolor{teal}{evening 6:45}. \newline
\textcolor{gray}{[System]} Please confirm that you need an appointment at Salon Revel at 6:45 pm later today. Allowed values ("Yes, go ahead","No").\newline
\textcolor{gray}{[User]} \textcolor{teal}{Yes}.
& Slot-51:[`Salon Revel'], \newline Slot-0:[`evening 6:45'], \newline Slot-154:[`the 1st'], \newline Slot-63:[`Yes, go \newline ahead'] \\
\hline
\textbf{Name splitting} \newline (tr:12296, v:2410, ts:10097) & 
Slot-76: Last name of the doctor\newline
Slot-161: Prefix name of the doctor\newline
Slot-75: Middle name of the doctor\newline
Slot-112: Appointment date with doctor\newline
Slot-31: Appointment time with doctor\newline
Slot-32: First name of the doctor
& 
\textcolor{gray}{[User]} I need to book a doctor's appointment for \textcolor{teal}{the 9th}. \newline 
\textcolor{gray}{[System]} Sure, what time, and do you have a preferred doctor \newline
\textcolor{gray}{[User]} Can you try for \textcolor{teal}{18:00} with \textcolor{teal}{dr. starks jayum bennett}? 
& Slot-76:[`bennett'], Slot-161:[`dr.'], Slot-75:[`jayum'], Slot-112:[`the 9th'], Slot-31:[`18:00'], Slot-32:[`starks'] \\
\hline
\textbf{ID data} \newline (tr:17860, v:2542, ts:4622) & 
Slot-19: City where the dentist is located \newline
Slot-54: id of the user
& 
\textcolor{gray}{[User]} Can you find dentist's listings? \newline
\textcolor{gray}{[System]} Can you give me your id. \newline
\textcolor{gray}{[User]} \textcolor{teal}{74563vQq} \newline
\textcolor{gray}{[System]} Do you have an area? \newline
\textcolor{gray}{[User]} I would like it in \textcolor{teal}{Mountain View}.
& Slot-19:[`Mountain View'], \newline Slot-54:[`74563vQq'] \\
\hline
\textbf{Address Parsing} \newline(tr:279, v:211, ts:160) & 
Slot-172: house-number \newline
Slot-108: street name \newline
Slot-168: name of the city/town/village \newline
Slot-157: state-district \newline
Slot-192: Number of riders. Allowed values ("1", "2", "3", "4")
& 
\textcolor{gray}{[User]} Can you call me at cab for \textcolor{teal}{one} person? I need to go to \textcolor{teal}{11 Hickson Road Walsh Bay}.
& Slot-172: [`11'], Slot-108: [`Hickson Road'], Slot-168: [`Walsh'], Slot-157: [`Bay'], Slot-192: [`1']
\\
\hline
\textbf{Relation} \newline (tr:12450, v:2523, ts:10395)& 
Slot-135: The amount of money to transfer\newline 
Slot-167: middle name \newline
Slot-212: first name \newline
Slot-202: last name \newline
Slot-214: prefix name \newline
Slot-145: relationship with receiver
& 
\textcolor{gray}{[User]} I'd like to make a \textcolor{teal}{\$370} transfer \newline 
\textcolor{gray}{[System]} Who do you want to send this money to? \newline
\textcolor{gray}{[User]} I want to send money to my \textcolor{teal}{brother George Sidney}
& 
Slot-135:[`\$370'], Slot-212:[`George'], Slot-202:[`Sidney'], Slot-145:[`brother'] \\
% \hline
% MD Format & 
% \begin{tabular}{p{4cm}} Slot-197: Speciality of the doctor. Allowed values ("Gynecologist", "ENT Specialist", "Ophthalmologist", "General Practitioner", "Dermatologist"). \newline
% Slot-15: City where the doctor is located.
% \end{tabular} 
% & 
% \begin{tabular}{p{6cm}}
% [USER] i am looking for a doctor. \newline
% [SYSTEM] Which **city**? Are you looking for Dermatologist or Ophthalmologist? \newline
% [USER] Looking for General Practitioner in Burlingame
% \end{tabular}  
% & Slot-15:['Burlingame'], Slot-197:['General Practitioner'] \\
\hline
\end{tabular}
\caption{Illustration of various Slot types. First column shows train(tr), val(v), and test(ts) sizes of each type. Second column shows slot library with random alphanumeric id and natural language description of slots. Slot filling model is expected to estimate values for these slots from the conversation shown in third column. Fourth column shows expected slot-filling model output. 
}
\label{tab:data-example}
\end{center}
\end{table*}

%\subsubsection{Categorical Slots}
%\subsection{Categorical Slots}

%Categorical slots come with set of permissible values from which the model has to map the value provided values to one of them. Model can only output one of those values for these slots. SGD dataset has many categorical slots to create samples which can help model to predict the categorical slots, but kind of variations we see for these slots in practical scenarios are not fully covered. For example, if the categorical values are of type ("yes","no") then model can get confused between user providing the values as yes or yes being said in confirming something as a casual conversation. SGD has boolean categorical slots that are similar to ("yes","no") and can be transformed to make data points to make model learn these kind of options. This data contains 17321 datapoints for train and 5058 for test set.

%\subsection{Slot Types}

%\vspace{-0.2cm}
~ \\
\textbf{4.1 SGD:} This part of data is prepared directly from SGD data by reshaping its original form into the format as described in Section \ref{sec:data-format}. It covers all the slot types that are part of the basic SGD slot library namely name, date, time, price, location, address, phone number, boolean value, etc. In this part, train size is 164982 and test size is 42297.

%price, location, city, date, time, address, boolean value, phone number, temperature, precipitation, humidity, wind

%\subsection{Multiple slots in single turn}
%Most of the dialogues in SGD data reveal slots as the conversation progresses and representation of data points where user comes in and types a long sentence revealing all the slots values for a particular intent are not there. This again is very common in practice where user can reveal all the slots in the very first utterance. To make model to see such scenarios, we took the confirmation messages coming from the SYSTEM utterance in SGD and paraphrased them into user tone using Falcon-180b and artificially created conversation data points that have multiple slots coming in a single turn. In this class, size of train and test set is 1056 and 286 respectively.
%\vspace{-0.2cm}
~ \\
\noindent
\textbf{4.2 Multiple Slots in Single Turn:} Most of the dialogues in SGD reveal slot values in a step-by-step manner over many turns. But revealing multiple slots in a single turn is a common scenario in industrial use cases such as banking, booking, etc. Hence to add that kind of data we used agent confirmation turns where typically many slot values given by the user are summarized in a single turn. We paraphrased these turns into user tone using \textit{Falcon-180b} \cite{falcon} and added to our data. %In this class, train size is 1056 and test 286.

%\vspace{-0.2cm}
~ \\
\noindent
\textbf{4.3 Long Slot Values:} Some of the slots in  industrial use cases can potentially take values ranging from a few words up to multiple sentences and even paragraphs. For example, issue description for tech support, preferences for hotel reservation, accident description for insurance claim, etc. Given the absence of such data in SGD, we manually created and added data simulating scenarios of hotel reservation, insurance claim, tech support, etc. 

%\vspace{-0.2cm}
~ \\
\noindent
\textbf{4.4 Categorical Slots:} These slots can take only a set of permissible values, hence the model is expected to map its output to one of the permissible values. Table \ref{tab:data-example} gives an illustration using a confirmation slot (i.e., Slot-63). % that can take one of the two values: \textit{Yes, go ahead} or \textit{No}. Any other value expressed by the user should get mapped to one of these two. 
Although SGD already includes data with categorical slots, we found a few limitations. For example, confirmation slots often get confused with the confirmation expressed elsewhere in the conversation but not related to the slot values. So we have augmented that part of the data with additional data prepared, also generated from SGD by transforming data corresponding to boolean categories into confirmations. %Data added in this manner amounts to a train size of 17321 and a test size of 5058.

%\subsection{Name splitting}
%Parsing Name into different components like, first name, middle name, last name etc is a very common scenario in industry use cases. SGD data captures name at a full name level wherever it occurs. To teach the model to learn address parsing, we took the data points coming from SGD with name and modified the slot library to have slots at finer granularity to cover for name parsing and appropriately split the full names into these components as part of the slot output and created sample data points. In this category, size of train and test set is 12296 and 10097 respectively. 
%\vspace{-0.2cm}
~ \\
\noindent
\textbf{4.5 Name Splitting:} Parsing names of people into first, middle and last names is a common requirement in industrial use cases. Although SGD contains name slots, do not have granular details. Hence to teach the model on name parsing, we have taken the name data from SGD and modified the slot library as well as annotations for finer granularity. %Train and test data sizes in this category are 12296 and 10097 respectively. 
Table \ref{tab:data-example} shows an example.

%\subsection{ID data}
%Practical slot filling scenarios usually have slots that capture IDs like employee number, user id, customer id etc. SGD data doesn't have any slots that fall into these categories. To address this gap, we introduced ID fields as part of the slot library for certain conversations and added system messages that probe for this slot and user message to capture these fields in various formats like numeric, alphanumeric, alphabets only etc. This data contains 17860 datapoints for train and 4622 for test set.
%\vspace{-0.2cm}
~ \\
\noindent
\textbf{4.6 ID Data:} Industrial use cases typically involve slots for different kinds of IDs such as employee number, user id, customer id etc. SGD do not have slots falling under these categories. To address this we introduced these ID fields as part of the slot library in certain conversations and added system messages to simulate probing for those slots and user message to simulate values in various formats like numeric, alphanumeric, alphabets only etc. %In this class, train size is 17860 and test size 4622.

%\subsection{Address Parsing}
%Parsing full address into different components like stree address, unit, city, state, country, zip code etc is another common requirement that we have seen in various use cases. For this again we created address parses for the existing addresses from SGD and introduced finer slots to capture address components as part of slot library and model output. This class of data has 279 and 160 train and test datapoints.
%\vspace{-0.2cm}
~ \\
\noindent
\textbf{4.7 Address Parsing:} Parsing full address into components like street address, unit, city, state, country, zip code etc is another common requirement in industrial use cases. For this again we took existing address data in SGD and reshaped with granular details. %Train size is 279 and test 160.

%\subsection{Relation Detection}
%Relationship of a person with another person is another common use case that comes in banking, insurance, health use cases very often. SGD data doesn't have slots that cater to capturing these attributes from the conversation. To address the gap we artificially introduced this slots in SGD data where some of the relationships are coming as part of user utterances and appropriately modified the slot output to be predicted. It has 12450 and 10395 datapoints for train and test sets.
%\vspace{-0.2cm}
~ \\
\noindent
\textbf{4.8 Relation:} Relationships among individuals is a common requirement in banking, insurance and health applications. SGD do not have any slot that we could reshape into this slot type. Hence we artificially added this slot into SGD data and annotated appropriately wherever relationships are expressed in the user utterances. %Train size is 12450 and test 10395.

%\subsection{Data with Markdown and other special formatting}
%Many dialogue engines use markdown or other special formatting when probing for certain slot values as part of their UI. When these prompts come as is to slot-filling model it can degrade its performance with the special characters present in the conversation. To counter this we artificially introduced markdown annotations around slot names when they are being probed for in the system utterance for certain sub-sample of SGD data.
%\subsubsection{Slot value overwrite} We don't have any special training for this so commenting. ***SUMIT*** Can we add a table with one sample for each of these types daa and train/dev/test sample sizes for each of these.

%\input{sections/realistic_benchmark}

\section{Experimental Evaluation}
\label{sec:results}

%In this section we present results on two different settings. One where we fine tune the models with data derived from SGD alone and second setting with full data consisting of various mixtures described in ~\ref{sec:data} on smaller models in with sizes less than 13b and show the importance of additional datasets that we curated for realistic slot-filling model. Table~\ref{tab:eval-results} shows the standard Macro F1 for various models on different test sets across 3 different models and we can see that model fine tuned with all the mixtures has better performance across different test sets.

\begin{table*}[t!]
\small
\centering
\begin{center}
\begin{tabular}{|r||c|c||c|c||c|c|}
\hline
\multicolumn{1}{|c||}{Test data} & \multicolumn{2}{c||}{\textit{Mistral} fine-tuned with} & \multicolumn{2}{c||}{\textit{Flan-T5} fine-tuned with} & \multicolumn{2}{c|}{\textit{granite.13b.v2} fine-tuned with} \\
\cline{2-7}
& SGD only & All data & SGD only & All data & SGD only & All data \\
\hline
SGD & 0.949 & \textbf{0.957} & \textbf{0.961} & \textbf{0.961} & 0.963 & \textbf{0.964} \\
Multiple Slots in Single Turn & 0.941 & \textbf{0.956} & 0.969 & \textbf{0.971} & 0.935 & \textbf{0.956} \\
Long Slot Values & 0.775 & \textbf{0.853} & 0.833 & \textbf{0.860} & 0.852 & \textbf{0.867} \\
% confirm & 0.915 & \textbf{0.946} & 0.928 & \textbf{0.953} & 0.948 & \textbf{0.957} \\
% category & 0.949 & \textbf{0.962} & 0.966 & \textbf{0.967} & 0.966 & \textbf{0.971} \\
Categorical Slots & 0.923 & \textbf{0.950} & 0.937 & \textbf{0.956} & 0.952 & \textbf{0.960} \\
Name Splitting & 0.804 & \textbf{0.926} & 0.813 & \textbf{0.932} & 0.808 & \textbf{0.940} \\
ID data & 0.949 & \textbf{0.965} & 0.899 & \textbf{0.970} & 0.954 & \textbf{0.972} \\
Address Parsing & 0.562 & \textbf{0.940} & 0.655 & \textbf{0.908} & 0.607 & \textbf{0.961} \\
Relation & 0.795 & \textbf{0.928} & 0.850 & \textbf{0.936} & 0.818 & \textbf{0.947} \\
\hline
Overall & 0.904 & \textbf{0.948} & 0.918 & \textbf{0.953} & 0.920 & \textbf{0.957} \\
\hline
\end{tabular}
\caption{Experimental evaluation of slot-filling model fine-tuned from three pre-trained LLMs: \textit{Mistral}, \textit{Flan-T5-XL} and \textit{granite.13b.v2}. Metric used is \text{Macro F1}, which is average of F1, a standard metric used.
%In the first column (for Test data) \textit{multi-slots} denote \textit{Single turn, multiple slots}. \textit{long slot} denote \textit{Long slot values}. 
%\textit{markdown} denote \textit{Markdown format}. 
%\textit{confirm} denote \textit{Confirmation slots}. \textit{category} denote \textit{Categorical slots}. \textit{name} denote \textit{Name splitting}. \textit{address} denote \textit{Address parsing}. 
%and
%\textit{relation} denote \textit{Relation extraction}.
%\textit{overwrite} denote \textit{Value overwrite}.
}
\label{tab:eval-results}
\end{center}
\end{table*}

In this section we report and analyze the results of experiments conducted to evaluate our prescribed approach for slot-filling model building.
Table \ref{tab:eval-results} shows results of experiments conducted using data described in Section \ref{sec:data} %\footnote{Table \ref{tab:data-example} in its first column gives train, validation and test splits of different datasets. Train and validation components are used for fine-tuning and test set is used for evaluation.} 
on three models as described in Section \ref{sec:fine-tuning}. These results are on two different settings aimed at measuring the usefulness of the additional data that we curated towards building useful slot-filling model from smaller-sized LLMs: 1) Using only SGD component of the data for fine-tuning and 2) Using entire (i.e., all) data for fine-tuning. Clearly the usage of entire data for fine-tuning is helping all 3 models to perform better on all individual slot types,
improving F1 metric by $\sim$4.2\% relative when averaged over all models and all datasets.

Note that the evaluation component of the  data (i.e., test data) described in Section \ref{sec:data} is actually held out from the data prepared for fine-tuning, in order to evaluate the performance of the models exactly on what it is made to learn. However, evaluation on data from a more realistic use-case would be representative of how the model is likely to perform during real-life usage in products. Hence we have also curated an additional test set named \textit{ Realistic Benchmark} from our in-house conversational assistants covering domains that our product is getting used. We describe evaluation on that next.

%of the However, it would be  the model and can only evaluate the slot-filling performance for the slot types  mainly to evalua

% \begin{itemize}
%     \item We train a model on SGD + the above datasets
%     \item We see significant improvement on the WA benchmark
%     \item We show that even small models (7B, 13B), finetuned on our dataset, can perform as well as very large models like ChatGPT.
%     \item Compare this model and the model trained on SGD only on the individual test sets created above. Model trained only on SGD lags behind.
%     \item Proves that the datasets that we created are necessary in order to get a robust, capable slot filling model that can handle the commonly occurring slots and conversation scenarios.
%     \item Our datasets and training methods are designed to incorporate zero shot capabilities in the slot filling models. Models can work out-of-the-box on new domains. 
% \end{itemize}
% \label{sec:results}

\if false
\begin{table*}[t!]
\small
\centering
\begin{center}
\begin{tabular}{|r|c|c|c|c|c|c||c|c|c|c|c|c|}
\hline
%\multicolumn{1}{|c|}{Test data} & \multicolumn{6}{c|}{Training data} \\
%\cline{2-13}
\multicolumn{1}{|c|}{Test data} & \multicolumn{6}{c||}{Mistral fine-tuned with} & \multicolumn{6}{c|}{Phi-2 fine-tuned with} \\
\cline{2-13}
& \multicolumn{3}{c|}{SGD only} & \multicolumn{3}{c||}{All data} & \multicolumn{3}{c|}{SGD only} & \multicolumn{3}{c|}{All data} \\
\cline{2-13}
& slot & slot & value & slot & slot & value & slot & slot & value & slot & slot & value \\
& key & value & match & key & value & match & key & value & match & key & value & match \\
& macro & f1 & score & macro & f1 & score & macro & f1 & score & macro & f1 & score \\
& f1 & & & f1 & & & f1 & & & f1 & & \\
\hline
%All & 0.952 & 0.917 & 0.885 & \textbf{0.974} & \textbf{0.950} & \textbf{0.903} & 0.942 & 0.892 & 0.855 & \textbf{0.958} & \textbf{0.923} & \textbf{0.881} \\
SGD & 0.975 & 0.949 & 0.898 & \textbf{0.979} & \textbf{0.957} & \textbf{0.905} & \textbf{0.979} & 0.963 & \textbf{0.917} & \textbf{0.979} & \textbf{0.964} & \textbf{0.917} \\
multi-slots & 0.949 & 0.941 & 0.983 & \textbf{0.963} & \textbf{0.956} & \textbf{0.987} & 0.947 & 0.935 & 0.973 & \textbf{0.963} & \textbf{0.956} & \textbf{0.985} \\
long slot & 0.926 & 0.775 & 0.674 & \textbf{0.969} & \textbf{0.853} & \textbf{0.746} & 0.971 & 0.852 & 0.741 & \textbf{0.972} & \textbf{0.867} & \textbf{0.765} \\
%markdown & \textbf{0.985} & 0.947 & 0.872 & 0.984 & \textbf{0.948} & \textbf{0.876} & \textbf{0.976} & 0.930 & 0.856 & 0.974 & \textbf{0.932} & \textbf{0.860} \\
confirm & 0.953 & 0.915 & 0.873 & \textbf{0.974} & \textbf{0.946} & \textbf{0.893} & 0.962 & 0.948 & \textbf{0.911} & \textbf{0.976} & \textbf{0.957} & 0.909 \\
category & 0.975 & 0.949 & 0.916 & \textbf{0.984} & \textbf{0.962} & \textbf{0.923} & 0.985 & 0.966 & 0.914 & \textbf{0.986} & \textbf{0.971} & \textbf{0.937} \\
name & 0.873 & 0.804 & 0.819 & \textbf{0.959} & \textbf{0.926} & \textbf{0.888} & 0.877 & 0.808 & 0.821 & \textbf{0.965} & \textbf{0.940} & \textbf{0.902} \\
ID data & 0.976 & 0.949 & 0.904 & \textbf{0.982} & \textbf{0.965} & \textbf{0.922} & 0.968 & 0.954 & 0.924 & \textbf{0.983} & \textbf{0.972} & \textbf{0.936} \\
address & 0.715 & 0.562 & 0.701 & \textbf{0.978} & \textbf{0.940} & \textbf{0.933} & 0.787 & 0.607 & 0.671 & \textbf{0.987} & \textbf{0.961} & \textbf{0.955} \\
relation & 0.862 & 0.795 & 0.826 & \textbf{0.960} & \textbf{0.928} & \textbf{0.894} & 0.870 & 0.818 & 0.854 & \textbf{0.969} & \textbf{0.947} & \textbf{0.908} \\
%overwrite & 0.955 & \textbf{0.924} & \textbf{0.945} & \textbf{0.956} & 0.923 & 0.944 & \textbf{0.951} & 0.907 & 0.922 & 0.949 & \textbf{0.907} & \textbf{0.925} \\
\hline
overall & 0.943 & 0.904 & 0.876 & \textbf{0.973} & \textbf{0.948} & \textbf{0.902} & 0.948 & 0.920 & 0.895 & \textbf{0.976} & \textbf{0.957} & \textbf{0.915} \\
\hline
\end{tabular}
\caption{Experimental evaluation of slot-filling model fine-tuned from two pre-trained LLMs: Mistral and Phi-2. In the first column (for Test data) \textit{multi-slots} denote \textit{Single turn, multiple slots}. \textit{long slot} denote \textit{Long slot values}. 
%\textit{markdown} denote \textit{Markdown format}. 
\textit{confirm} denote \textit{Confirmation slots}. \textit{category} denote \textit{Categorical slots}. \textit{name} denote \textit{Name splitting}. \textit{address} denote \textit{Address parsing}. 
and
\textit{relation} denote \textit{Relation extraction}.
%\textit{overwrite} denote \textit{Value overwrite}.
}
\label{tab:eval-results}
\end{center}
\end{table*}
\fi

\if false
\begin{table*}[t!]
\small
\centering
\begin{center}
\begin{tabular}{|r|c|c|c|c|c|c||c|c|c|c|c|c|}
\hline
%\multicolumn{1}{|c|}{Test data} & \multicolumn{6}{c|}{Training data} \\
%\cline{2-13}
\multicolumn{1}{|c|}{Test data} & \multicolumn{6}{c||}{Mistral fine-tuned with} & \multicolumn{6}{c|}{Phi-2 fine-tuned with} \\
\cline{2-13}
& \multicolumn{3}{c|}{SGD only} & \multicolumn{3}{c||}{All data} & \multicolumn{3}{c|}{SGD only} & \multicolumn{3}{c|}{All data} \\
\cline{2-13}
& slot & slot & value & slot & slot & value & slot & slot & value & slot & slot & value \\
& key & value & match & key & value & match & key & value & match & key & value & match \\
& macro & f1 & score & macro & f1 & score & macro & f1 & score & macro & f1 & score \\
& f1 & & & f1 & & & f1 & & & f1 & & \\
\hline
%All & 0.952 & 0.917 & 0.885 & \textbf{0.974} & \textbf{0.950} & \textbf{0.903} & 0.942 & 0.892 & 0.855 & \textbf{0.958} & \textbf{0.923} & \textbf{0.881} \\
SGD & 0.975 & 0.949 & 0.898 & \textbf{0.979} & \textbf{0.957} & \textbf{0.905} & \textbf{0.974} & \textbf{0.943} & 0.886 & 0.968 & 0.938 & \textbf{0.889} \\
multi-slots & 0.949 & 0.941 & 0.983 & \textbf{0.963} & \textbf{0.956} & \textbf{0.987} & \textbf{0.939} & \textbf{0.919} & 0.943 & 0.894 & 0.885 & \textbf{0.969} \\
long slot & 0.926 & 0.775 & 0.674 & \textbf{0.969} & \textbf{0.853} & \textbf{0.746} & 0.923 & 0.702 & 0.582 & \textbf{0.926} & \textbf{0.789} & \textbf{0.692} \\
%markdown & \textbf{0.985} & 0.947 & 0.872 & 0.984 & \textbf{0.948} & \textbf{0.876} & \textbf{0.976} & 0.930 & 0.856 & 0.974 & \textbf{0.932} & \textbf{0.860} \\
confirm & 0.953 & 0.915 & 0.873 & \textbf{0.974} & \textbf{0.946} & \textbf{0.893} & 0.953 & 0.916 & 0.870 & \textbf{0.963} & \textbf{0.933} & \textbf{0.882} \\
category & 0.975 & 0.949 & 0.916 & \textbf{0.984} & \textbf{0.962} & \textbf{0.923} & 0.975 & 0.936 & 0.872 & \textbf{0.976} & \textbf{0.938} & \textbf{0.876} \\
name & 0.873 & 0.804 & 0.819 & \textbf{0.959} & \textbf{0.926} & \textbf{0.888} & 0.867 & 0.784 & 0.788 & \textbf{0.931} & \textbf{0.885} & \textbf{0.862} \\
ID data & 0.976 & 0.949 & 0.904 & \textbf{0.982} & \textbf{0.965} & \textbf{0.922} & \textbf{0.977} & 0.876 & 0.794 & 0.975 & \textbf{0.947} & \textbf{0.899} \\
address & 0.715 & 0.562 & 0.701 & \textbf{0.978} & \textbf{0.940} & \textbf{0.933} & 0.720 & 0.511 & 0.603 & \textbf{0.908} & \textbf{0.791} & \textbf{0.797} \\
relation & 0.862 & 0.795 & 0.826 & \textbf{0.960} & \textbf{0.928} & \textbf{0.894} & 0.860 & 0.784 & 0.805 & \textbf{0.936} & \textbf{0.888} & \textbf{0.864} \\
%overwrite & 0.955 & \textbf{0.924} & \textbf{0.945} & \textbf{0.956} & 0.923 & 0.944 & \textbf{0.951} & 0.907 & 0.922 & 0.949 & \textbf{0.907} & \textbf{0.925} \\
\hline
overall & 0.943 & 0.904 & 0.876 & \textbf{0.973} & \textbf{0.948} & \textbf{0.902} & 0.942 & 0.892 & 0.855 & \textbf{0.958} & \textbf{0.923} & \textbf{0.881} \\
\hline
\end{tabular}
\caption{Experimental evaluation of slot-filling model fine-tuned from two pre-trained LLMs: Mistral and Phi-2. In the first column (for Test data) \textit{multi-slots} denote \textit{Single turn, multiple slots}. \textit{long slot} denote \textit{Long slot values}. 
%\textit{markdown} denote \textit{Markdown format}. 
\textit{confirm} denote \textit{Confirmation slots}. \textit{category} denote \textit{Categorical slots}. \textit{name} denote \textit{Name splitting}. \textit{address} denote \textit{Address parsing}. 
and
\textit{relation} denote \textit{Relation extraction}.
%\textit{overwrite} denote \textit{Value overwrite}.
}
\label{tab:eval-results}
\end{center}
\end{table*}
\fi

\if false
\begin{table*}[t!]
\small
\centering
\begin{center}
\begin{tabular}{|r|c|c|c|c|c|c||c|c|c|c|c|c|}
\hline
%\multicolumn{1}{|c|}{Test data} & \multicolumn{6}{c|}{Training data} \\
%\cline{2-13}
\multicolumn{1}{|c|}{Test data} & \multicolumn{6}{c||}{Mistral fine-tuned with} & \multicolumn{6}{c|}{Phi-2 fine-tuned with} \\
\cline{2-13}
& \multicolumn{3}{c|}{SGD only} & \multicolumn{3}{c||}{All data} & \multicolumn{3}{c|}{SGD only} & \multicolumn{3}{c|}{All data} \\
\cline{2-13}
& slot & slot & value & slot & slot & value & slot & slot & value & slot & slot & value \\
& key & value & match & key & value & match & key & value & match & key & value & match \\
& macro & f1 & score & macro & f1 & score & macro & f1 & score & macro & f1 & score \\
& f1 & & & f1 & & & f1 & & & f1 & & \\
\hline
%All & 0.952 & 0.917 & 0.885 & \textbf{0.974} & \textbf{0.950} & \textbf{0.903} & 0.942 & 0.892 & 0.855 & \textbf{0.958} & \textbf{0.923} & \textbf{0.881} \\
SGD & \textbf{0.981} & 0.959 & 0.\textbf{907} & 0.981 & \textbf{0.959} & 0.906 & \textbf{0.974} & \textbf{0.943} & 0.886 & 0.968 & 0.938 & \textbf{0.889} \\
multi-slots & 0.929 & 0.916 & 0.969 & \textbf{0.938} & \textbf{0.931} & \textbf{0.979} & \textbf{0.939} & \textbf{0.919} & 0.943 & 0.894 & 0.885 & \textbf{0.969} \\
long slot & 0.955 & 0.751 & 0.608 & \textbf{0.970} & \textbf{0.874} & \textbf{0.777} & 0.923 & 0.702 & 0.582 & \textbf{0.926} & \textbf{0.789} & \textbf{0.692} \\
%markdown & \textbf{0.985} & 0.947 & 0.872 & 0.984 & \textbf{0.948} & \textbf{0.876} & \textbf{0.976} & 0.930 & 0.856 & 0.974 & \textbf{0.932} & \textbf{0.860} \\
confirm & 0.963 & 0.933 & 0.885 & \textbf{0.980} & \textbf{0.956} & \textbf{0.895} & 0.953 & 0.916 & 0.870 & \textbf{0.963} & \textbf{0.933} & \textbf{0.882} \\
category & 0.985 & 0.959 & \textbf{0.902} & \textbf{0.988} & \textbf{0.961} & 0.899 & 0.975 & 0.936 & 0.872 & \textbf{0.976} & \textbf{0.938} & \textbf{0.876} \\
name & 0.887 & 0.819 & 0.819 & \textbf{0.957} & \textbf{0.929} & \textbf{0.895} & 0.877 & 0.784 & 0.788 & \textbf{0.931} & \textbf{0.885} & \textbf{0.862} \\
ID data & 0.985 & 0.965 & 0.916 & \textbf{0.987} & \textbf{0.968} & \textbf{0.916} & \textbf{0.977} & 0.876 & 0.794 & 0.975 & \textbf{0.947} & \textbf{0.899} \\
address & 0.695 & 0.517 & 0.650 & \textbf{0.985} & \textbf{0.951} & \textbf{0.941} & 0.720 & 0.511 & 0.603 & \textbf{0.908} & \textbf{0.791} & \textbf{0.797} \\
relation & 0.872 & 0.816 & 0.843 & \textbf{0.957} & \textbf{0.928} & \textbf{0.897} & 0.860 & 0.784 & 0.805 & \textbf{0.936} & \textbf{0.888} & \textbf{0.864} \\
%overwrite & 0.955 & \textbf{0.924} & \textbf{0.945} & \textbf{0.956} & 0.923 & 0.944 & \textbf{0.951} & 0.907 & 0.922 & 0.949 & \textbf{0.907} & \textbf{0.925} \\
\hline
overall & 0.952 & 0.917 & 0.885 & \textbf{0.974} & \textbf{0.950} & \textbf{0.903} & 0.942 & 0.892 & 0.855 & \textbf{0.958} & \textbf{0.923} & \textbf{0.881} \\
\hline
\end{tabular}
\caption{Experimental evaluation of slot-filling model fine-tuned from two pre-trained LLMs: Mistral and Phi-2. In the first column (for Test data) \textit{multi-slots} denote \textit{Single turn, multiple slots}. \textit{long slot} denote \textit{Long slot values}. 
%\textit{markdown} denote \textit{Markdown format}. 
\textit{confirm} denote \textit{Confirmation slots}. \textit{category} denote \textit{Categorical slots}. \textit{name} denote \textit{Name splitting}. \textit{address} denote \textit{Address parsing}. 
and
\textit{relation} denote \textit{Relation extraction}.
%\textit{overwrite} denote \textit{Value overwrite}.
}
\label{tab:eval-results}
\end{center}
\end{table*}
\fi
\subsection{A Realistic Benchmark for Slot-Filling}
\label{sec:template_data}
%We also looked at the in house conversation assistant and curated some commonly used use cases based test cases that cover across different domains where the product is being used. 
Over the years, our product \textit{WatsonX Assiatant}~\cite{watsonxasst} built a collection of pre-built templates that could be used to create conversation agents. These pre-built templates capture realistic use cases across different domains. These templates contain both slots expected for a given use case along with the conversation flow guiding the slot-filling and overall conversation. Following an approach similar to Multiwoz, a human annotator will create an agent using one of these templates and interact with the agent to generate the conversation and slot annotation.  We curated about 354 conversation samples across domains like healthcare, insurance, banking, HR etc. This dataset has a flavour of various kinds of scenarios we talked about in Section~\ref{sec:data}. 

We bench-marked this test set against three fine tuned models as discussed Section \ref{sec:fine-tuning} (and reported on in Table \ref{tab:eval-results}) and also against a few popular LLMs through prompting, such as ChatGPT3.5 Turbo~\cite{chatgpt-turbo}, Azure GPT3.5 Turbo~\cite{azure}, PALM/text-bison-001~\cite{palm} and llama-2-70b-chat\cite{touvron2023llama}. For prompting, we took the best prompt that was built using a small sample data for all the prompted models. Results shown in Table~\ref{tab:real-eval-results} reveal that with task-specific fine tuning we are able to achieve better performance even with smaller-sized fine tuned models, with additional benefit of lower latency. Our \textit{granite.13b.v2} model has achieved relative improvements of $\sim$6.9\% and $\sim$57\% against the best performing prompted model on F1 metric and average latency, respectively.

%using all the data from previous section and also on the popular external language models though prompting. We took the best prompt that was built using a small sample data for all the prompted Models. Table~\ref{tab:real-eval-results} shows the numbers on ChatGPT3.5 Turbo~\cite{chatgpt-turbo}, Azure GPT3.5 Turbo~\cite{azure}, PALM/text-bison-001~\cite{palm} and llama-2-70b-chat\cite{touvron2023llama}. We see that with task specific fine tuning we can achieve very good performance with the smaller fine tuned models. 

\begin{table}[t!]
\small
\centering
\begin{center}
\begin{tabular}{|r|c|c|c|}
\hline
Model & Method  & Macro F1 &  Latency \\
\hline
ChatGPT3.5 Turbo & Prompted & 0.87 & 2.82 \\
Azure gpt-35-turbo & Prompted & 0.80 & 4.16 \\
PALM/text-bison-001 & Prompted & 0.79 & 1.77 \\
llama-2-70b-chat & Prompted & 0.70 & 6.27 \\
Flan-T5-XL & Full FT  & 0.90 & 1.43 \\
Mistral-7b & LoRA FT & 0.85 & 0.65 \\
granite.13b.v2 & LoRA FT & \textbf{0.93} & 0.75 \\
\hline
\end{tabular}
\caption{Experimental evaluation of various models on realistic benchmark. All the models have Decoder only architecture except for \textit{Flan-T5-XL} which has Encoder-Decoder architecture. Latency is average time taken for one slot filling request. FT refers to Fine Tuning.
}
\vspace{-0.2cm}
\label{tab:real-eval-results}
\vspace{-0.5cm}
\end{center}
\end{table}

% \begin{itemize}
%     \item Method used for creating the test sets
%     \item Number of data points
%     \item Types of slots and conversation scenarios covered - long slot value, name split, address parsing, ID, confirmation slots, etc
% \end{itemize}

% \subsection{Baselines}
% \begin{itemize}
%     \item Describe the procedure used to evaluate ChatGPT, PaLM, etc on this benchmark.
%     \item Describe how we trained Mistral, Phi-2 on SGD. How did they perform? 
%     \item Should we justify why we only trained on SGD and not MultiWOZ, or other Slot filling datasets out there?
%     \item Create a table with the leader-board entries and refer to it here.
%     \item Smaller LMs with 7-13B params, trained only on SGD, don't perform well on the realistic WA benchmark. This shows the need for richer training data (the datasets that we created)
% \end{itemize}

\vspace{-0.2cm}
\section{Conclusions}
\vspace{-0.2cm}
In this paper, we described methodology we used to build an industry-grade slot-filling model to work with conversations assistants. We used readily available dialogue state tracking dataset SGD along with a few derived and custom curated datasets to fine tune smaller-sized LLMs, that in turn are able to achieve state of the art performance for slot-filling. We demonstrated this using benchmarks derived as held-out during training data creation and a realistic benchmark gathered from the in-house conversational assistants used in products. Method we used to fine tune the model results in a zero shot slot filling model that can work across different domains using slot descriptions as clue to gather slot values.

\bibliography{main}

\appendix

\section{Appendix}
\begin{figure}[t!]
\begin{lstlisting}[language=json,firstnumber=1]
Input Prompt:

Find all the slots and their values from conversation. 

<slot library>
Slot-211: first name
Slot-847: last name
Slot-615: email
Slot-196: add phone number. Allowed values ("Yes", "No")
Slot-240: phone number
Slot-437: date of birth - MM/DD/YYYY
Slot-568: confirm correct. Allowed values ("Yes", "No")

<conversation>
[USER] I'd like to register
[SYSTEM] I can help you with that! To create an account, we will need a few pieces of information. Let me guide you through this process! First, please provide your first name. 
[USER] My first name is Jim
[SYSTEM] Thanks! Now, enter your last name. 
[USER] Fullman
[SYSTEM] Awesome. I'll also need an email address associated with your new account. We will need to send you an email to confirm it's really you!
Which email address would you like to use? 
[USER] Please use jfullman@bitterdrink.com
[SYSTEM] For added security, we can also add a phone number to your account. Would you like to do so? option: ("Yes", "No")
[USER] Yes
[SYSTEM] Okay, please enter your phone number below. 
[USER] 1233211234
[SYSTEM] Finally, I'll also need your date of birth in MMDDYYYY format. This will also serve as your 8-digit authentication PIN.
Keep this number handy in case you are locked out of your account. 
[USER] 10011999
[SYSTEM] Please enter a valid date of birth in MM/DD/YYYY format. 
[USER] 10/01/1999
[SYSTEM] Thanks! Does the following information look correct? If so, I will proceed with creating your account.
  - Name: Fullman, Jim
  - Email Address: jfullman@bitterdrink.com
  - Phone Number: 1233211234
  - Birthday: Oct 1, 1999
option ("Yes","No")
[USER] Yes
[SYSTEM] You did it! Your new account is being created and you will receive an email confirmation shortly.
Welcome! 
[USER] I will login now

Output:
'Slot-211': 'Jim',
'Slot-847': 'Fullman'
'Slot-615': 'jfullman@bitterdrink.com',
'Slot-196': 'Yes',
'Slot-240': '1233211234',
'Slot-437': '10/01/1999',
'Slot-568': 'Yes',
\end{lstlisting}
    \caption{Sample Input Prompt and Output for User Registration use case.}
    \label{fig:prompt1}
\end{figure}

Here we present sample prompts based on realistic benchmark explained in Section~\ref{sec:template_data} through figures (Figure~\ref{fig:prompt1} to \ref{fig:prompt3}). These figures contain input prompt followed by output produced by \textit{granite.13b.v2} model. Figure~\ref{fig:prompt1} shows prompt for user registration use case. Prompt starts with one line instruction followed by slot library and then conversation. Slot library starts with \texttt{<slot library>} tag. Slot library defines all slots that are supposed to be filled from conversation. For each slot in the library, slot id is of form \texttt{"Slot-<random number>"}. Slot id is followed by natural language description of the slot. 
\begin{figure}[h!]
\begin{lstlisting}[language=json,firstnumber=1]
Input Prompt:

Find all the slots and their values from conversation. 

<slot library>
Slot-412: Which account would you like to pull the funds from? Allowed values ("Checking account", "Savings account")
Slot-581: How much of the bill would you like to pay? 
Slot-314: confirm correctness. Allowed values ("Yes", "No")

<conversation>
[USER] I want to Pay my remaining money
[SYSTEM] Which account would you like to pull the funds from?
option ("Checking account", "Savings account")
[USER] Savings account
[SYSTEM] How much of the bill would you like to pay? 
[USER] 125
[SYSTEM] Let me confirm, you'd like to pay $125.00 from your Savings account. Is that correct?
option ("Yes","No")
[USER] Yes

Output:
'Slot-412': 'Savings account',
'Slot-581': '125',
'Slot-314': 'Yes'
\end{lstlisting}
    \caption{Sample Input Prompt and Output for Money transfer use case.}
    \label{fig:prompt2}
\end{figure}

\begin{figure}[h!]
\begin{lstlisting}[language=json,firstnumber=1]
Input Prompt:

Find all the slots and their values from conversation. 

<slot library>
Slot-758: How can we best contact you? Allowed values ("by email", "by phone")
Slot-770: Great. What's your phone number? 
Slot-116: And finally, can you describe your problem in more detail? 

<conversation>
[USER] Something's broken
[SYSTEM] Sorry to hear that. Let's get a support ticket created. How can we best contact you?
option ("by email", "by phone")
[USER] by phone
[SYSTEM] Great. What's your phone number? 
[USER] 4561231234
[SYSTEM] And finally, can you describe your problem in more detail? I'll use this information to submit a ticket to our team. 
[USER] I am locked out of my account

Output:
'Slot-758': 'by phone',
'Slot-770': '4561231234',
'Slot-116': 'I am locked out of my account'
\end{lstlisting}
    \caption{Sample Input Prompt and Output for Support use case.}
    \label{fig:prompt3}
\end{figure}
Next, under \texttt{<conversation>} tag, User and System turns of the conversation are given wherein utterances are preceded by tags \texttt{[USER]} or \texttt{[SYSTEM]}. Based on this conversation, the model is expected to estimate the slot values at its output. 

Figures~\ref{fig:prompt2} and \ref{fig:prompt3} show prompts for money transfer and support ticket creation use cases. All these use cases illustrate a variety of slots including name splitting, category, long slot values etc. 

\end{document}